\documentclass[conference]{IEEEtran}
\usepackage{times}
\usepackage{epsfig}
\usepackage{subcaption}
\usepackage{graphicx}
\usepackage{amsmath}
\usepackage{amssymb}
\usepackage{float}
\usepackage{array, makecell, cellspace}
\usepackage{multirow}
\usepackage{balance}
\usepackage[font=small,skip=5pt]{caption}
\def\bs{\ensuremath\boldsymbol}

\usepackage[pagebackref=true,breaklinks=true,colorlinks,bookmarks=false]{hyperref}

\begin{document}

\title{Guided Generative Models using Weak Supervision for Detecting Object Spatial Arrangement in Overhead Images}
\makeatletter
\newcommand{\linebreakand}{%
  \end{@IEEEauthorhalign}
  \hfill\mbox{}\par
  \mbox{}\hfill\begin{@IEEEauthorhalign}
}
\makeatother

\author{Weiwei Duan\\
University of Southern California\\
{\tt\small weiweiduan@usc.edu}

\and
Yao-Yi Chiang\\
University of Minnesota\\
{\tt\small yaoyi@umn.edu}

\and
Stefan Leyk\\
University of Colorado Boulder\\
{\tt\small stefan.leyk@colorado.edu }
\linebreakand
\and
Johannes H. Uhl\\
University of Colorado Boulder\\
{\tt\small johannes.uhl@colorado.edu}

\and
Craig A. Knoblock\\
University of Southern California\\
{\tt\small knoblock@isi.edu}
}

\maketitle

\begin{abstract}
The increasing availability and accessibility of numerous overhead images allows us to estimate and assess the spatial arrangement of groups of geospatial target objects, which can benefit many applications, such as traffic monitoring and agricultural monitoring. Spatial arrangement estimation is the process of identifying the areas which contain the desired objects in overhead images. Traditional supervised object detection approaches can estimate accurate spatial arrangement but require large amounts of bounding box annotations. Recent semi-supervised clustering approaches can reduce manual labeling but still require annotations for all object categories in the image. This paper presents the target-guided generative model (TGGM), under the Variational Auto-encoder (VAE) framework, which uses Gaussian Mixture Models (GMM) to estimate the distributions of both hidden and decoder variables in VAE. Modeling both hidden and decoder variables by GMM reduces the required manual annotations significantly for spatial arrangement estimation. Unlike existing approaches that the training process can only update the GMM as a whole in the optimization iterations (e.g., a "minibatch"), TGGM allows the update of individual GMM components separately in the same optimization iteration. Optimizing GMM components separately allows TGGM to exploit the semantic relationships in spatial data and requires only a few labels to initiate and guide the generative process. Our experiments shows that TGGM achieves results comparable to the state-of-the-art semi-supervised methods and outperformes unsupervised methods by 10\% based on the $F_{1}$ scores, while requiring significantly fewer labeled data.

\end{abstract}

\section{Introduction}
Overhead images provide data to unlock information unseen from the ground. Detecting locations for groups of similar-sized geospatial objects of interest (i.e., target objects) is an important computer vision task, which can benefit various applications, such as detecting a group of cars to monitor busy hours in parking lots~\cite{traffic:1,traffic:2}, surveying crowds for disease monitoring~\cite{covid19}, and detecting a group of trees in plantations for agricultural monitoring purposes~\cite{ob_geo:plantations1,ob_geo:plantations2}. We define detecting the location of a group of target objects as estimating the spatial arrangement of target objects in this paper. The state-of-the-art supervised object detectors~\cite{ssd,yolov3,faster_rcnn} can estimate the spatial arrangement of target object locations by obtaining a precise location of each target object but require thousands of bounding-box level annotations for the specific objects. However, overhead images, such as satellite imagery and scanned topographic maps, do not have sufficient bounding-box level annotations to train supervised object detectors. Although many public scenic image datasets, such as PASCAL Visual Object Classes (VOC) Challenge~\cite{pascal-voc-2012}, provide bounding-box level annotations, transfer learning technologies~\cite{transfer_learning,iccv2017:wsob_tl,iccv2019:wsob6_tl} cannot effectively transfer knowledge from scenic images to overhead images because annotations for scenic images are typically for large and notable objects, while overhead images often contain small and densely clustered or dispersed objects over a large area. The spatial arrangement estimation for a group of similar-sized target objects, such as cars, does not require precise location for each target object. Therefore, the bounding-box level annotations are not necessary. 

Annotating a region of interest (ROI) and a few boxes covering target objects within the ROI is sufficient to detect a group of target objects in overhead images. ROI is a region covering a group of target objects in overhead images. For example, most of the overlapping green boxes at the bottom right panel in Figure~\ref{fig:workflow} represent areas with a significant overlap with one or more cars. These boxes, typically generated using a sliding-window-based method, can be used to estimate the spatial arrangement of car objects in a parking lot. Therefore, a sliding-window-based method is often sufficient to estimate the spatial arrangement of a group of objects within an ROI.

A sliding-window-based method uses a fix-size window sliding across an ROI and detects the windows that contain the target objects within the ROI. The group of detected windows containing the target objects is assumed to be representive of the spatial arrangement of the target objects. In this case, the sliding-window-based method aims to group window areas in an ROI into two categories: target vs. non-target categories (e.g., cars vs. other objects in a parking lot). The solutions to this type of clustering task can be either unsupervised~\cite{VaDE,dual_ae,GMVAE} or semi-supervised methods~\cite{m1_m2,AVAE,biva}.

Unsupervised clustering methods~\cite{dual_ae,VaDE} discover the separable patterns in an ROI to separate windows into clusters. However, the target-vs.-non-target pattern might not be the only and dominant separable pattern within an ROI. An unsupervised clustering method could use other image patterns (e.g., lighting or texture) to separate the windows. Hence, results from unsupervised methods are not robust for various types of target objects and ROIs (i.e., the target objects might not be grouped into one cluster). In contrast, semi-supervised models~\cite{AVAE,biva, m1_m2} can separate the windows into a specific target and non-target category by using annotated areas of example objects (labeled windows containing target and non-target objects) to guide the clustering process. However, these methods typically require substantial manual work of labeling hundreds of target and non-target examples. 

This paper proposes a target-guided generative model (TGGM), which exploits a few labeled target windows to detect the spatial arrangement of the target objects within an ROI in overhead images. TGGM has two main advantages: 1. TGGM only needs labeled windows for the target category instead of all categories required by semi-supervised methods; 2. TGGM reduces the number of required labeled target windows from hundreds in the existing approaches to just a few. 
TGGM exploits a few labeled target windows to initialize the target cluster and then iteratively optimizes the clustering process by accumulating windows assigned to the target cluster. This way, the target cluster can eventually cover all target objects with diverse appearances within an ROI. TGGM also leverages a unique property in spatial data. The ROI boundaries are carefully selected to have a strong semantic relationship with the target objects (e.g., parking lots and cars). Also, within the ROI, the non-target objects are similar (e.g., most non-car areas in a parking lot are parking lot surfaces). If the input image is geo-referenced, TGGM can use external geographic data sources (e.g., OpenStreetMap\footnote{\url{https://www.openstreetmap.org/}}) to automatically generate the desired ROI boundaries (e.g., parking lot boundaries). In other cases, annotating the ROI boundary is a straightforward task compared to annotating a large number of bounding boxes for objects within the ROI. 

\begin{figure}[ht]
 	\centering
 	\includegraphics[width=0.45\textwidth]{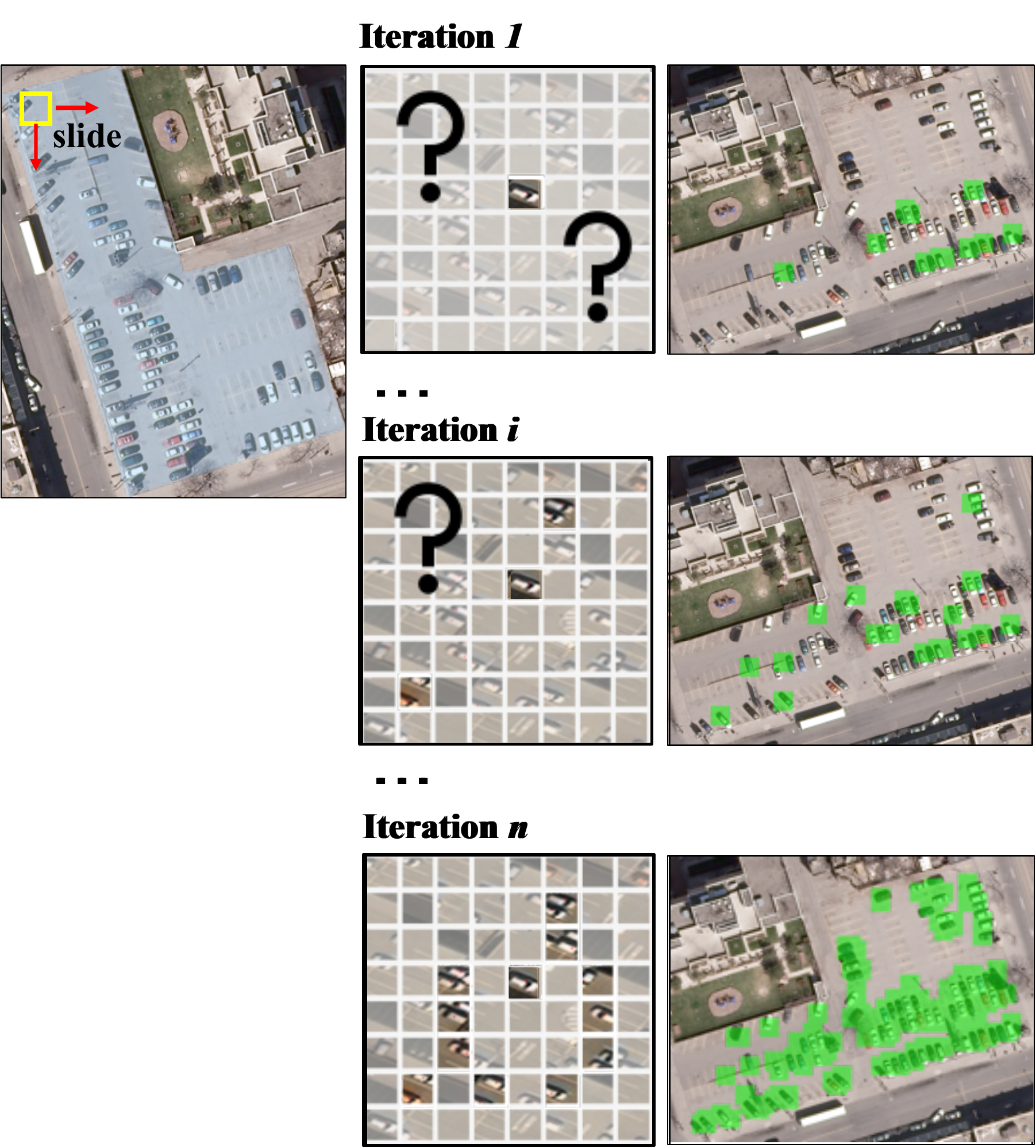}
 	\caption{\small The first column shows that TGGM uses a window (similar to the size of a car) sliding across the blue ROI. The second column shows the generated windows across the ROI. The non-blurred and blurred images in the second column represent labeled target windows and unlabeled windows, respectively. In the first iteration, the non-blurred image is manually annotated. TGGM takes the manually labeled target window and unlabeled windows as inputs first and detects more windows covering cars in the green areas in the top panel in the third column. The detected target windows are inputs for the next iteration to help detect new target windows. The iteration ends when TGGM  cannot detect new target windows. In the end, the green areas in the bottom left panel show the detected target windows representing the spatial arrangement of cars in the parking lot.}
 	\label{fig:workflow}
 \end{figure}

Figure~\ref{fig:workflow} illustrates how TGGM detects the target windows (e.g., windows overlapping with one ore more cars) within the ROI iteratively. The first column shows how to generate fixed-size windows over the blue ROI. The window has approximately the same size as the target objects (e.g., cars in the ROI). The second column shows the generated windows across the ROI. The non-blurry windows in the second column indicate the initial manually labeled target windows (in iteration 1) or the detected target windows (in the rest of iterations). The green areas in the third column represent the detected windows containing cars in the parking lot. TGGM leverages an iterative detection process that takes advantage of the detected target windows for the next iteration. The iteration ends when TGGM cannot detect new target windows. The bottom right panel in Figure~\ref{fig:workflow} shows that TGGM estimates the spatial arrangement of cars in the parking lot.
 
TGGM is a weakly-supervised probabilistic clustering framework based on the Variational Auto-encoder and Gaussian Mixture Models. The main contribution of TGGM is that it can exploit a few labeled data to guide the clustering process so that the data in one of the resulting clusters are similar to the labeled data.  Specifically, unlike the existing approaches in which the training process can only update the Mixture of Gaussians (MoG) as a whole or use a subset of the data to update a specific component in the MoG in separate optimization iterations (e.g., a "minibatch"), TGGM's network design allows the update of individual components in the MoG separately in the same optimization iteration. Using labeled and unlabeled data in one optimization iteration is important because when the data in some categories are (partially) labeled, an optimization iteration should update the MoG using both labeled and unlabeled data so that the data distributions of some specific components in MoG is specific for the partially labeled data in some categories. 

Without loss of generality, here, we assume 1) the input data contain two clusters: the target and the non-target clusters, and 2) only limited labeled data in the target category are available. TGGM assumes the hidden space is an MoG where the number of components is the number of the desired clusters (e.g., two components in this case) like the other generative clustering approaches~\cite{VaDE}. Unlike other approaches, TGGM can use both the labeled and unlabeled data in one optimization iteration. Initially, the labeled data provide examples for the target cluster, and the unlabeled data can contain either the target or non-target data. In one optimization iteration, for the labeled and unlabeled data in the target cluster from the previous iteration, TGGM only updates the parameters for the target component in the MoG. For all other unlabeled data, TGGM updates the entire MoG. The result is that TGGM iteratively learns the MoG parameters that best describe the target data in one component and all other data in the other component. We call this process the target guidance mechanism.

In sum, TGGM exploits a few manually labeled target windows in a spatially-constrained ROI to iteratively detect the spatial arrangement of the target objects within the ROI in overhead images. This paper's contribution is that TGGM provides the optimization procedure that leverages both labeled target and unlabeled windows in one optimization iteration. The optimization procedure enables using the labeled target windows and unlabeled windows to guide the formation of the target and non-target clusters. The resulting windows in the target cluster represent the spatial arrangement of the target objects in the overhead images.
\section{Related Work}

Weakly supervised object detection~\cite{iccv2019:wsob1_instance,iccv2019:wsob2,iccv2019:wsob3,iccv2019:wsob4,iccv2019:wsob5,iccv2019:wsob6_tl,iccv2017:wsob_tl,iccv2017:wsob1,iccv2017:wsob2,iccv2017:wsob3,iccv2015:wsob}, aiming to localize objects with image-level annotations, has attracted attention because bounding box annotations require intensive manual work. However, existing weakly supervised object detectors focus on scenic images, such as images in the PASCAL VOC Challenge~\cite{pascal-voc-2012}. The target objects, such as cats, are usually large and notable in scenic images. Because of the notability property and proportional size of target objects in scenic images, existing methods transfer prior knowledge learned from scenic images~\cite{iccv2017:wsob_tl,iccv2019:wsob6_tl} or utilize the multiple instances learning methods~\cite{iccv2019:wsob1_instance} to localize the objects using image annotations. However, the notability property of target objects is usually not valid in overhead images~\cite{iccv2019:ob_geo,ob_geo:slide_win2}. The target objects in satellite imagery, like cars, are small, densely clustered or spatially dispersed over large areas~\cite{ob_geo:slide_win3}. Therefore, the weakly supervised object detectors cannot directly apply on overhead images. 

Sliding-window-based detectors are commonly used to detect if a window covers small target objects in overhead images~\cite{ob_geo:slide_win1,ob_geo:slide_win2,ob_geo:slide_win3,ob_geo:survey}, because The state-of-the-art object detectors~\cite{yolov3,faster_rcnn} lose the information about small objects when taking entire satellite imagery as the input because of down-sampling operations. The sliding window is a straightforward way to search for small objects across imagery. Detecting if the sliding window contains the target objects is equivalent to classifying the images into the target or non-target categories. State-of-the-art image classifiers without or with a limited number of labeled images are unsupervised and semi-supervised methods.

Unsupervised methods separate data into multiple clusters by identifying separable patterns in the data. Generative models under the Variational Auto-encoder (VAE) framework are common unsupervised clustering methods~\cite{GMVAE,VaDE,biva,CaGeM,dual_ae}, which learn the distribution to represent and separate inputs in the hidden space. Generative clustering models have shown impressive results by learning flexible distribution representations in the hidden space. However, images in sliding windows have many separable patterns, such as light-vs.-dark or target-vs.-non-target patterns. The methods do not produce target and non-target clusters when unsupervised methods use other separable patterns to cluster the images. In contrast, the target guidance mechanism guides TGGM to separate images in sliding windows into the target and non-target clusters. 

Semi-supervised learning methods~\cite{biomedical,biva,m1_m2,social_images,constraints_detector,weak_supervised:biomedical,weak_supervised:saliency,weak_supervised:GAN,semi_supervised:image_retrieval,semi_supervised:VAE,semi_supervised:VAE_2017} aim to automatically annotate large amounts of unlabeled data using a small set of labeled data in each category. Recent work~\cite{biva,m1_m2} shows that unsupervised generative clustering models can convert to semi-supervised models easily by adding a cross-entropy loss for labeled data. The cross-entropy loss optimized by labeled data helps to improve the clustering results. However, labeling both target and non-target windows can require a huge amount of manual work. In contrast, TGGM reduces the manual work for non-target window annotations and separates windows into the target and non-target categories by leveraging the strong semantic relationship between the ROI and the target objects.

\section{Target-Guided Generative Model}
First, we introduce symbols used in the following subsections. $\boldsymbol{x}_{t}$ represents labeled target windows. $\boldsymbol{x}_{u}$ is unlabeled windows. $\boldsymbol{\hat{x}}_{u}$ stands for the generated unlabeled windows from TGGM. $\boldsymbol{z}$ is a continuous variable to represent the distribution of $\boldsymbol{x}_{t}$ and $\boldsymbol{x}_{u}$ in the hidden space. $y$ is a categorical variable representing the labels for $\boldsymbol{x}_{u}$. $y=\{0,1\}$, 1 for target windows and 0 for non-target windows. 

Figure~\ref{fig:network} shows TGGM's architecture. $f_{inf}$ encodes $\boldsymbol{x_{u}}$ concatenating with $y$ into $\boldsymbol{z}$, then $f_{gen}$ decodes $\boldsymbol{z}$ into $\boldsymbol{\hat{x}_{u}}$. $\boldsymbol{\hat{x}_{u}}$ is the reconstructed $\boldsymbol{x_{u}}$. $f_{cls}$ assigns $\boldsymbol{x_{u}}$ into clusters. Section 3.1 describes the distribution of $\boldsymbol{z}$ given $\boldsymbol{x_{u}}$ and $y$ learned by $f_{inf}$ and $f_{cls}$. Section 3.2 describes the process for $f_{prior}$. Section 3.3 formulates the evidence lower bounds of the marginal likelihood of $\boldsymbol{x_{u}}$ and $\boldsymbol{x_{t}}$ to optimize TGGM. Section 3.4 explains how TGGM leverages labeled target windows to form target and non-target clusters. 

\begin{figure}[ht]
	\centering
	\includegraphics[width=0.45\textwidth]{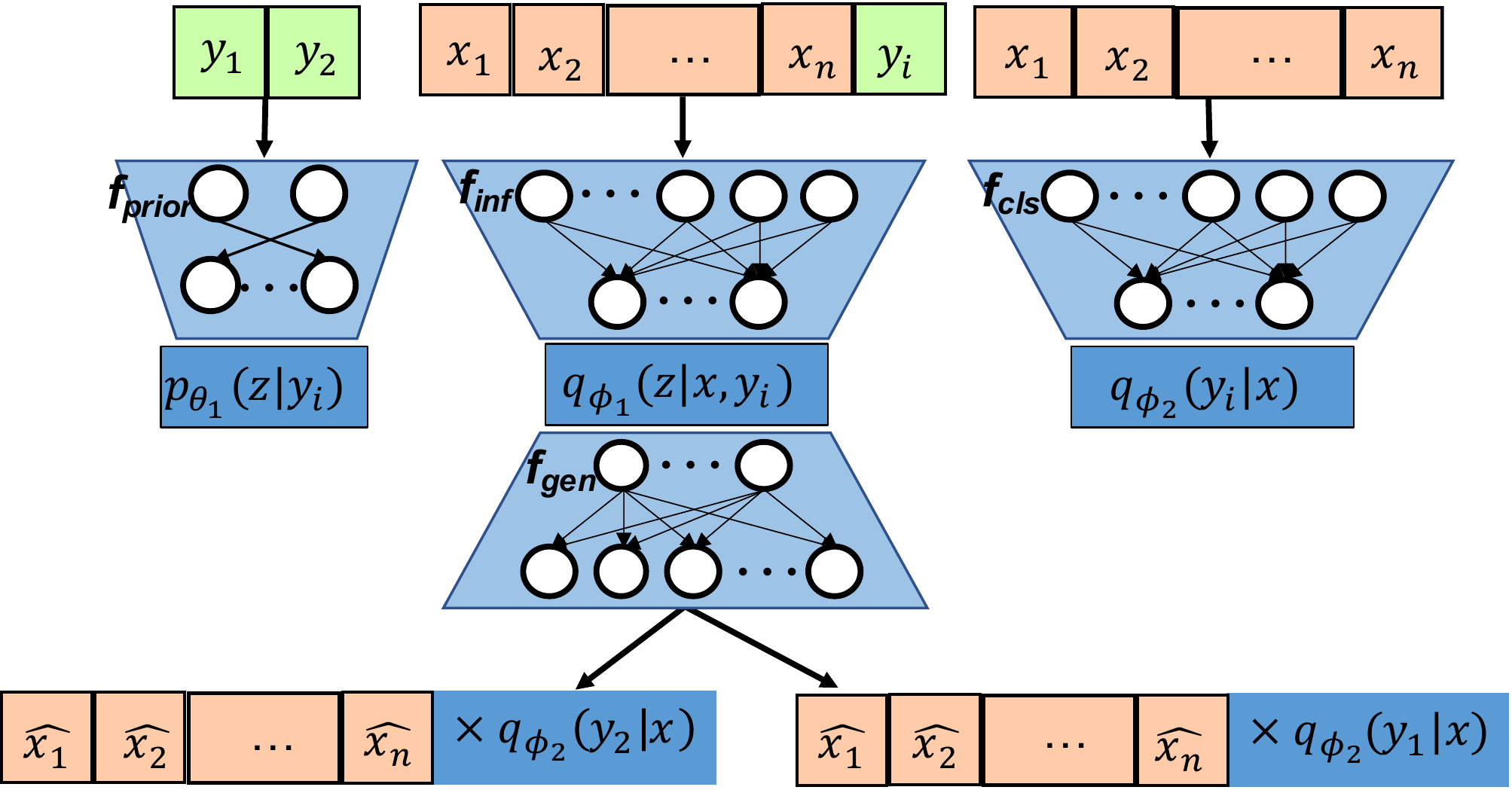}
	\caption{\small TGGM's architecture. $f_{inf}$ in the center model encodes $\boldsymbol{x_{u}}$ concatenating with $y$ into the distribution of the hidden variable $z$, and $f_{gen}$ decodes $\boldsymbol{z}$ into $\boldsymbol{\hat{x}_{u}}$ weighted by $q(y|\boldsymbol{x_{u}})$ from $f_{cls}$ represented by the right model. Section 3.1 describes $f_{inf}$ and $f_{cls}$. Section 3.2 describes $f_{prior}$ represented by the left model. Sections 3.3 and 3.4 describe the optimization procedure.} 
	\label{fig:network}
\end{figure}

\subsection{The Variational Inference}
TGGM learns the approximation posterior distribution for $\boldsymbol{z}$, i.e., $q(\boldsymbol{z}|\boldsymbol{x}_{u})$ to estimate the true posterior probability $p(\boldsymbol{z}|\boldsymbol{x}_{u})$ which is intractable~\cite{vae}. The approximate posterior probability of $\boldsymbol{z}$ is MoG with two Gaussian components showing in eq. ~\ref{eq:posterior_z}. The parameters of Gaussian components are learned by $f_{inf}$ in eq.~\ref{eq:z_f}. The weights of two components are the probabilities of the hidden categorical variable $y$ in eq.~\ref{eq:posterior_y}. The hidden categorical variable $y$ is learned by $f_{cls}$ activated by the softmax function. The joint approximate posterior probability of $\boldsymbol{z}$ and $y$ is shown in eq.~\ref{eq:inference_u}.
\begin{flalign}\label{eq:posterior_z}
&	q(\boldsymbol{z}|\boldsymbol{x}_{u})=\sum_{y}q(y|\boldsymbol{x}_{u})q(\boldsymbol{z}|\boldsymbol{x}_{u},y)
\end{flalign}
\begin{flalign}\label{eq:z_f}
	q(\boldsymbol{z}|\boldsymbol{x}_{u},y) = \mathcal{N}(\tilde{\mu_{\boldsymbol{z}}}, \tilde{\sigma_{\boldsymbol{z}}}), \hspace{0.3cm} [\tilde{\mu_{\boldsymbol{z}}}, \tilde{\sigma_{\boldsymbol{z}}}] = f_{inf}(\boldsymbol{x}_{u},y)
\end{flalign}
\begin{flalign}\label{eq:posterior_y}
	q(y|\boldsymbol{x}_{u})=Cat(y|\pi(\boldsymbol{x}_{u}))=softmax(f_{cls}(\boldsymbol{x}_{u}))
\end{flalign}
\begin{flalign}\label{eq:inference_u}
	q(\boldsymbol{z},y|\boldsymbol{x}_{u}) = q(\boldsymbol{z}|\boldsymbol{x}_{u},y)q(y|\boldsymbol{x}_{u})
\end{flalign}
\subsection{The Generative Process for Windows}
Generating a target window follows the steps below:
\begin{enumerate}
	\item Sample a hidden vector $\boldsymbol{z}$ from $p(\boldsymbol{z}|y=1)$ 
	\item Compute the distribution of $\boldsymbol{x}_{t}$ following $\mathcal{N}(\mu_{x_{t}}, \sigma_{x_{t}})$
	\item Sample a $\boldsymbol{x}$ from $\mathcal{N}(\mu_{x_{t}}, \sigma_{x_{t}})$
\end{enumerate}

The generative process for non-target windows is similar to the above process, except sampling from $p(\boldsymbol{z}|y=0)$ and sampling $\boldsymbol{x}$ from the Gaussian distribution for non-target windows. $p(\boldsymbol{z}|y)$ in step 1 above follows a Gaussian distribution showing in eq.~\ref{eq:generate_z}. The parameters of Gaussian distribution are learned by $f_{prior}(y)$ in eq.~\ref{eq:generate_z_param}. The parameters of $\mathcal{N}(\mu_{x_{t}}, \sigma_{x_{t}})$ in step 2 are learned by $f_{gen}$ in eq.~\ref{eq:generate_xt_param}.
\begin{align}\label{eq:generate_z} 
 p(\boldsymbol{z}|y)=\mathcal{N}(\mu_{\boldsymbol{z}}, \sigma_{\boldsymbol{z}}) 
\end{align}
\begin{align}\label{eq:generate_z_param} 
	[\mu_{\boldsymbol{z}}, \sigma_{\boldsymbol{z}}] = f_{prior}(y)
\end{align}
\begin{align}\label{eq:generate_xt_param} 
[\mu_{x_{t}},\sigma_{x_{t}}] = f_{gen}(\boldsymbol{z})
\end{align}
According to the generative process above, the joint probability $p(\boldsymbol{x},\boldsymbol{z},y)$ can be factorized as:
\begin{flalign}\label{eq:generate_joint}
	 p(\boldsymbol{x},\boldsymbol{z},y) = p(\boldsymbol{x}|\boldsymbol{z},y)p(\boldsymbol{z}|y)p(y)
\end{flalign}
Where $p(y)$ is the prior distribution for $y$ defined in eq.~\ref{eq:generate_y}.
\begin{align}\label{eq:generate_y}
 	p(y)=Cat(1/K)
\end{align}
Where $K$ is the number of categories. In our case, the number of categories is two, one for the target category and the other for the non-target one. At the beginning of the iterations, TGGM assumes that the number of target and non-target windows, i.e., $p(y)$ is uniformly distributed since we do not have prior knowledge about the number of target objects in the ROI. This assumption is relaxed in later iterations.

\subsection{The Generative and Inference Processes for\\ Labeled Windows}
For the labeled target windows, $y$ becomes an observation instead of a hidden variable, i.e., $y=1$. Because $y$ is an observation, TGGM only uses labeled target windows to learn one Gaussian component of $\boldsymbol{z}$ and generate windows from the component. As a result, the approximate posterior distribution of $\boldsymbol{z}$ and generative $\boldsymbol{x}_{t}$ distribution can be written as eq.~\ref{eq:generative_zt} and eq.~\ref{eq:generative_xt}, respectively.
\begin{flalign}\label{eq:generative_zt}
	q(\boldsymbol{z} | \boldsymbol{x}_{t})=q(\boldsymbol{z}|\boldsymbol{x}_{t},y=1)
\end{flalign}
\begin{flalign}\label{eq:generative_xt}
	& p(\boldsymbol{x}_{t}|\boldsymbol{z},y=1) = p(\boldsymbol{x}_{t}|\boldsymbol{z},y=1)p(\boldsymbol{z}|y=1)
\end{flalign}

\subsection{Evidence Lower Bounds}
TGGM aims to separate target and non-target windows into two clusters by maximizing the marginal likelihood of unlabeled windows, $\boldsymbol{x}_{u}$. Because the log-likelihood of $\boldsymbol{x}_{u}$ (the leftmost term in eq.~\ref{eq:likehood_x}) is intractable, TGGM optimizes the evidence lower bound ($\mathcal{L}_{ELBO}$) (the rightmost term in eq.~\ref{eq:likehood_x}) instead.
\begin{align}\label{eq:likehood_x}
	&\log p(\boldsymbol{x}_{u}) = \log\sum_{y}\int_{\boldsymbol{z}} p(\boldsymbol{x}_{u},\boldsymbol{z},y)d\boldsymbol{z} \nonumber\\
	&\geq \mathbb{E}_{q(\boldsymbol{z},y|\boldsymbol{x}_{u})}\Big[\log \frac{p(\boldsymbol{x}_{u},\boldsymbol{z},y)}{q(\boldsymbol{z},y|\boldsymbol{x}_{u})}\Big] = \mathcal{L}_{ELBO}(\boldsymbol{x}_{u})      
\end{align}
Eq.~\ref{eq:elbo_u} shows the details of $\mathcal{L}_{ELBO}$ for unlabeled windows. $\mathcal{L}_{ELBO}$ after using the reparameterization trick proposed in VAE~\cite{vae} can be written as:
\begin{align}\label{eq:elbo_u}
	\mathcal{L}_{ELBO}(\boldsymbol{x}_{u})
	&=     \mathbb{E}_{q(\boldsymbol{z},y|\boldsymbol{x}_{u})}\Big[\log \frac{p(\boldsymbol{x}_{u},\boldsymbol{z},y)}{q(\boldsymbol{z},y|\boldsymbol{x}_{u})}\Big] \nonumber \\
	&=\sum_{y}q(y|\boldsymbol{x}_{u})\log p(\boldsymbol{x_{u}}|\boldsymbol{z},y) \nonumber \\
	&\phantom{=}-\sum_{y}q(y|\boldsymbol{x}_{u})\mathcal{KL}\Big[q(\boldsymbol{z}|\boldsymbol{x}_{u},y)\|p(\boldsymbol{z}|y)\Big] \nonumber \\
	&\phantom{=}-\mathcal{KL}\Big[q(y|\boldsymbol{x}_{u})\|p(y)\Big]
\end{align}
According to the generative and variational inference processes for labeled target windows, $\boldsymbol{x}_{t}$, described in the last subsection, TGGM uses $\boldsymbol{x}_{t}$ to optimize one Gaussian component in $\boldsymbol{z}$ and generate $\boldsymbol{x}_{t}$ from the component. $\mathcal{L}_{ELBO}$ for $\boldsymbol{x}_{t}$ is defined in eq.~\ref{eq:elbo_l}.
\begin{align}\label{eq:elbo_l}
	\mathcal{L}_{ELBO} (\boldsymbol{x}_{t})
	& = \mathbb{E}_{q(\boldsymbol{z}|\boldsymbol{x}_{t},y=1)}\Big[\log \frac{p(\boldsymbol{x}_{t},\boldsymbol{z},y=1)}{q(\boldsymbol{z}|\boldsymbol{x}_{t},y=1)}\Big] \nonumber\\
	& = \log p(\boldsymbol{x_{t}}|\boldsymbol{z},y=1)\nonumber\\
	&\phantom{=}- \mathcal{KL}\Big[q(\boldsymbol{z}|\boldsymbol{x}_{t},y=1)\| p(\boldsymbol{z}|y=1)\Big]
\end{align}
The total $\mathcal{L}_{ELBO}$ as the optimization goal of TGGM is the summation of $\mathcal{L}_{ELBO}$ for $\boldsymbol{x}_{u}$ and $\boldsymbol{x}_{t}$ showing in eq.~\ref{eq:elbo}.
\begin{align}\label{eq:elbo}
	\mathcal{L}_{ELBO}(\boldsymbol{x})=\mathcal{L}_{ELBO}(\boldsymbol{x}_{u})+\mathcal{L}_{ELBO}(\bs{x}_{t})
\end{align}

\subsection{Network Design}
Figure~\ref{fig:network} shows TGGM's network architecture. When the input is a labeled target window $\boldsymbol{x}_{t}$, the encoder ($f_{inf}$) in TGGM encodes $\boldsymbol{x}_{t}$ concatenating with $y=1$ into $q(\boldsymbol{z}|\boldsymbol{x}_{u},y=1)$, then decoder ($f_{gen}$) sampled $\boldsymbol{z}$ from $p(\boldsymbol{z}|y=1)$ into $\hat{\boldsymbol{x_{t}}}$. In the optimization process, the optimization goal of the second term in eq.~\ref{eq:elbo_l} is that one component of the hidden variable $\boldsymbol{z}$ represents the labeled target windows. The optimization goal of the first term in eq.~\ref{eq:elbo_l} is that the decoder generates similar $\hat{\boldsymbol{x_{t}}}$ to $\boldsymbol{x}_{t}$ from the component of $\boldsymbol{z}$ which represent labeled target windows in the hidden space. TGGM explicitly learns one component in $\boldsymbol{z}$ to represent the labeled target windows. We call this learning process the target guidance mechanism.


When the input is a unlabeled window, $\boldsymbol{x}_{u}$, $f_{inf}$ encodes $\boldsymbol{x}_{u}$ concatenating with $y$ into the hidden variable $\boldsymbol{z}$. $f_{gen}$ decodes the sampled $\boldsymbol{z_{0}}$ from $p(\boldsymbol{z}|y=0)$ and $\boldsymbol{z_{1}}$ from $p(\boldsymbol{z}|y=1)$ into $\boldsymbol{\hat{x}}_{u0}$ and $\boldsymbol{\hat{x}}_{u1}$. In the optimization process, the optimization goal of first term in eq.~\ref{eq:elbo_u} is assigning a high weight ($q(y|\boldsymbol{x}_{u})$) to $\boldsymbol{\hat{x}}_{u0}$ or $\boldsymbol{\hat{x}}_{u1}$, whichever is more similar to $\boldsymbol{x}_{u}$. For example, when the input is an unlabeled window covering the target objects, the $\boldsymbol{\hat{x}}_{u1}$ from $p(\boldsymbol{z}|y=1)$ should be more similar to $\boldsymbol{\hat{x}}_{u}$ than $\boldsymbol{\hat{x}}_{u0}$ from $p(\boldsymbol{z}|y=0)$ because $p(\boldsymbol{z}|y=1)$, optimized by labeled target windows in the target guidance mechanism, represents the target windows. The weight for $\boldsymbol{\hat{x}}_{u1}$ is higher than the weight for $\boldsymbol{\hat{x}}_{u0}$ (i.e., ($q(y=1|\boldsymbol{x}_{u}) > q(y=0|\boldsymbol{x}_{u})$)

Similarly, the optimization goal of the second term in eq.~\ref{eq:elbo_u} is assigning a high weight ($q(y|\boldsymbol{x}_{u})$) to one component of $\boldsymbol{z}$, which has the smaller \textit{KL} value than the other component does. For example, when the input is an unlabeled window covering the target object, the \textit{KL} value between $q(\boldsymbol{z}|\boldsymbol{x}_{u},y=1)$ and $p(\boldsymbol{z}|y=1)$ should be smaller than the \textit{KL} value between $q(\boldsymbol{z}|\boldsymbol{x}_{u},y=0)$ and $p(\boldsymbol{z}|y=0)$ (i.e. $\mathcal{KL}[q(\boldsymbol{z}|\boldsymbol{x}_{u},y=1)\| p(\boldsymbol{z}|y=1)]<\mathcal{KL}[q(\boldsymbol{z}|\boldsymbol{x}_{u},y=0)\| p(\boldsymbol{z}|y=0)]$) because $q(\boldsymbol{z}|\boldsymbol{x}_{u},y=1)$ and $p(\boldsymbol{z}|y=1)$, optimized by labeled target windows in the target guidance mechanism, represent the target windows. Hence, TGGM assigns a higher weight to the component in $\boldsymbol{z}$ which represents the target windows (i.e., ($q(y=1|\boldsymbol{x}_{u}) > q(y=0|\boldsymbol{x}_{u})$). The weights, $q(y|\boldsymbol{x_{u}})$, are the probabilities of a window belonging to the target and non-target clusters. With the target guidance mechanism, TGGM assigns unlabeled windows covering the target objects into the target cluster. 

\section{Experiment and Analysis}
This section presents a comprehensive experiment on TGGM with four types of target objects in four datasets, which are cars, airplanes, and ships in overhead imagery~\cite{dior,COWC,xview}, and wetland symbols from scanned topographic maps. The experiment entails two groups of experiments and sensitivity analysis. The first group of experiments applied TGGM on four types of objects in five datasets, and evaluated the spatial arrangement estimation from TGGM by comparing with three baseline models. The second group of experiments applied TGGM on the DIOR dataset and compared results from TGGM with a supervised object detector, YOLOv3~\cite{yolov3}.


\subsection{Data Preparation and Experimental Settings}
\textbf{Datasets and Target Objects}\hspace{0.4cm}We tested the TGGM on four objects in two types of images in four datasets. They are cars in the Cars Overhead With Context (COWC) dataset~\cite{COWC}, cars and airplanes in the xView dataset~\cite{xview}, and airplanes and ships in the DIOR dataset~\cite{dior}, which are all overhead imagery. The other object is wetland symbols in scanned topographic maps from the United States Geological Survey (USGS) topographic map archive.

\textbf{Region-Level Annotations}\hspace{0.4cm} We used two approaches to generate the ROI annotations for the two groups of experiments, respectively. In the first group of experiments, we manually drew polygons which have an assumed strong semantic relationship with the target objects in overhead imagery in the xView, COWC, and DIOR datasets. The USGS dataset in the first set of experiments was combined with an external dataset\footnote{https://apps.nationalmap.gov/downloader} to provide the ROIs for wetland symbols. In the second group of experiments, we tested on the entire imagery covering a group of ships and airplanes in the DIOR dataset.

\textbf{Target Window Annotations}\hspace{0.4cm} We manually annotate one window covering a target object for each ROI. To augment the number of labeled target windows, we rotated the window and translated the positions of the target object in the window. 

\textbf{Evaluation Methods}\hspace{0.4cm}  In the first group of experiments, we used precision, recall, and $F_{1}$ at the grid-cells to estimate the performance of spatial arrangement estimation using TGGM. We sliced images into non-overlapped grid cells. Grid-level assessments are commonly used in Geospatial Information Science (GISc) to estimate the detection accuracy~\cite{grid-level_est:1,grid-level_est:2,grid-level_est:3}. In the second group of experiments, we used mean average precision (mAP). In the first group of experiments, we generated the grid-level ground truth from the bounding-box ground truth. We sliced images using varied grid sizes, i.e., $20\times20$-, $40\times40$-, $60\times60$-, $80\times80$-pixel. If the overlapping area between the grid cell and the bounding box of the target object is over either 50\% bounding-box or 50\% grid cell, the grid cell was a true positive. Otherwise, it was a true negative. xView, COWC, and DIOR datasets provide the bounding box annotations for target objects. We manually generated the bounding boxes ground truth for wetland symbols in the USGS dataset. In the second group of experiments, we used the bounding-box ground truth provided by the DIOR dataset. We generated grid-level results from the detected windows. If the overlapping area between the grid and detected target window occupied over either 50\% window or 50\% grid, the grid was a detected positive grid. Otherwise, it was a detected negative grid. In the second group of experiments, the DIOR dataset provides the bounding boxes ground truth. We generate bounding-box results from detected windows using non-max suppression (NMS)~\cite{faster_rcnn}.

\textbf{Baselines}\hspace{0.4cm} In the first group of experiments, we compared TGGM with two unsupervised generative clustering models, VaDE~\cite{VaDE} and dualAE~\cite{dual_ae}, and one semi-supervised generative model, AVAE~\cite{AVAE}. For the semi-supervised model (AVAE), we used 40\% target and non-target windows as labeled data for the training phase. The number of labeled windows for AVAE follows recently reported experiments~\cite{AVAE}. In the second group of experiments, we compered TGGM with YOLOv3, a supervised object detector. We adopted the YOLOv3 results from paper~\cite{dior}, which trained YOLOv3 in a supervised manner.

\textbf{Implementation Details}\hspace{0.4cm} All submodels in TGGM were the multilayer perceptrons with two fully connected layers and optimized by the Adam optimizer with a learning rate of $1\mathrm{e}{-3}$. The iterative learning of TGGM for all tasks ended around seven iterations. Each iteration converged around 200 epochs. 

\subsection{Experiment Results and Analysis}


Figure~\ref{fig:cmp_chart} shows that TGGM outperformed the unsupervised generative clustering models, i.e., dualAE and VaDE, and performed similar to the semi-supervised generative model, i.e., AVAE, in the first group of experiments. An $F_{1}$ score that is on average 10\% higher than for dualAE and VaDE demonstrates that the spatial arrangement estimation using TGGM was more accurate than the state-of-the-art unsupervised methods. The high recall and low precision from the unsupervised methods show that the clustering results are noisier than those from TGGM. The precise results from TGGM show that the target guidance mechanism improves the clustering results. Compared with the semi-supervised model, AVAE, the $F_{1}$ score for TGGM was about 5\% lower on average. However, AVAE required 40\% labeled target and non-target windows, while TGGM only needed one labeled target window with augmentation, which minimized the manual work needed. 

Figure~\ref{fig:xView_car} shows the spatial arrangement estimation of cars from TGGM and three baselines in xView. The green and red boxes are true positive and false positive grids, respectively. The bottom two figures show the noisy results from unsupervised models. Although TGGM missed some true positive grids compared to results from AVAE (the semi-supervised method), the spatial arrangement from TGGM covers most cars in the parking lot. In sum, TGGM could obtain more accurate spatial arrangement estimation within the ROI than unsupervised models did and achieved similar results as semi-supervised with much less manual work. 

\begin{figure}[h]
	\centering
	\includegraphics[width=0.4\textwidth]{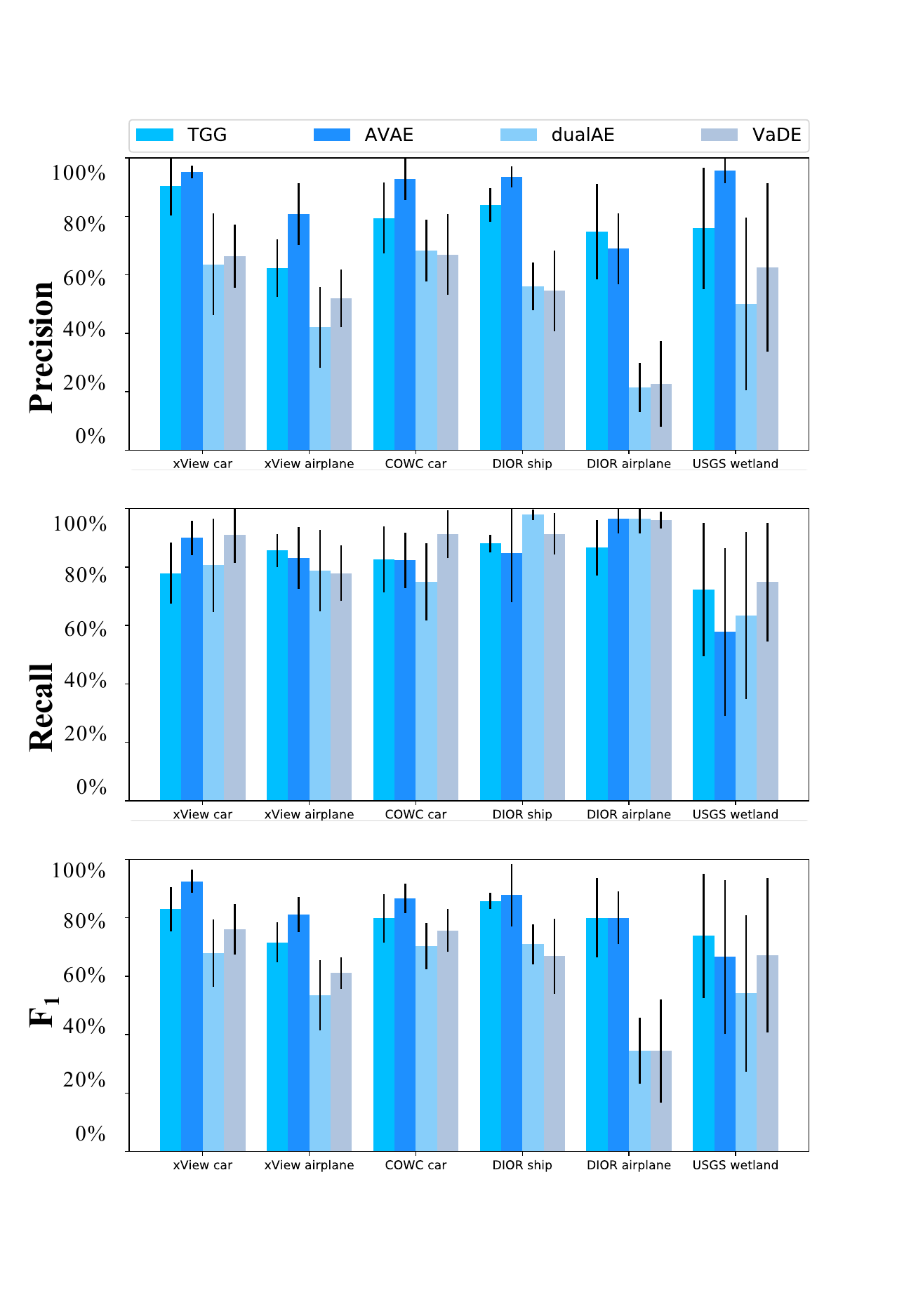}
	\caption{\small The performance comparisons among TGGM and three baseline models. VaDE and dualAE are unsupervised baselines, and AVAE is the semi-supervised baseline.}
	\label{fig:cmp_chart}
\end{figure}

\begin{figure}[ht]
	\centering
	\includegraphics[width=0.4\textwidth]{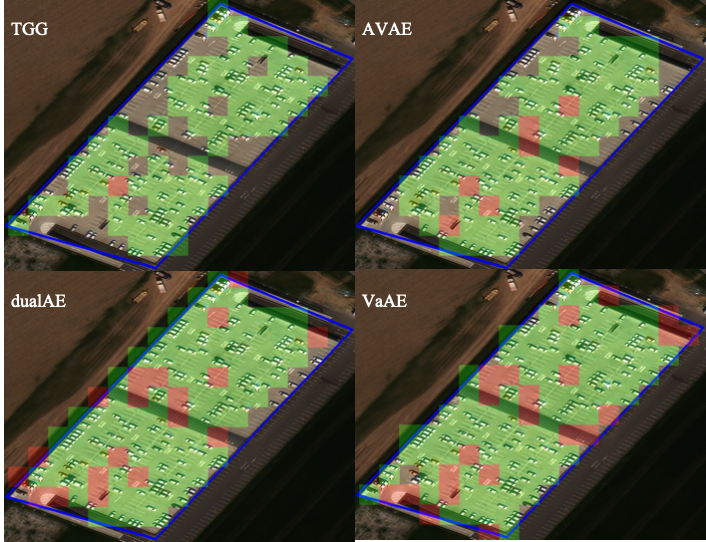}
	\caption{\small The evaluation of car spatial arrangement estimation using $40\times40$-pixel grid size. Green and red boxes represent true positive and false positive grids, respectively.}
	\label{fig:xView_car}
\end{figure}

In the second group of experiments, we tested the detection of ships and airplanes in the DIOR dataset and compared the results with the best supervised model, YOLOv3~\cite{dior} using mAP. The results from YOLOv3 represent the best case scenario where a large amount of training data for the target objects are available. For airplanes, TGGM obtained 60.15\% mAP, while YOLOV3 achieved 72.2\%. As for the ships, the mAP for TGGM was 69.92\%, while mAP from YOLOv3 was 87.4\%. The average mAP from TGGM was $15\%$ lower than YOLOv3, but TGGM only required one labeled target window with data augmentation for each image while YOLOv3 needed $50\%$ target objects for the training process. 

Two major reasons cause mAP from TGGM lower than YOLOv3. First, the ROI annotations may not have a strong semantic relationship with the target objects. We chose all imagery covering a group of target objects in the DIOR dataset. This criterion cannot guarantee any semantic relationship between the imagery and the target objects. Figure~\ref{fig:DIOR_failure1} shows that the imagery covers not only the airplane apron, which has a strong semantic relationship with airplanes but also other areas, such as a lawn. The weak semantic relationship between airplanes and the imagery contents results in rather noisy results (Figure~\ref{fig:DIOR_failure1}). Figure~\ref{fig:DIOR_failure2} shows similar failures where the image covers a large area, much of which has no particular semantic relationship with ships. When we manually chose the ROI which has strong semantic relationship with the target objects in the first grounp of experiment, TGGM achieved $77\%$ average $F_{1}$ score.

The other reason is the limited performance of TGGM on multi-scaled objects. TGGM uses a fixed-size window slide across all imagery and separates windows into target and non-target categories. The sizes of target objects varied a lot in all imagery. The fixed-size window, which is set based on the size of the small target objects, would often be smaller than those large target objects, and thus the model would fail to cluster the large target objects into the target category. Figure~\ref{fig:DIOR_failure3} shows that TGGM could detect small ships but missed some of the large ships when using small target objects to set the window size. However, users can adjust the window size to fit both large and small target objects when applying TGGM to one ROI. Consequently, TGGM can estimate the spatial arrangement of the target objects reliably. Figures~\ref{fig:DIOR_success1},  ~\ref{fig:DIOR_success2}, and ~\ref{fig:USGS_success3} show that TGGM applying to specific ROIs performs well in estimating the spatial arrangements of airplanes and ships in the overhead imagery, and wetland symbols in the topographic map, respectively.

\begin{figure}[ht]
	\begin{subfigure}{.15\textwidth}
		\includegraphics[width=\linewidth]{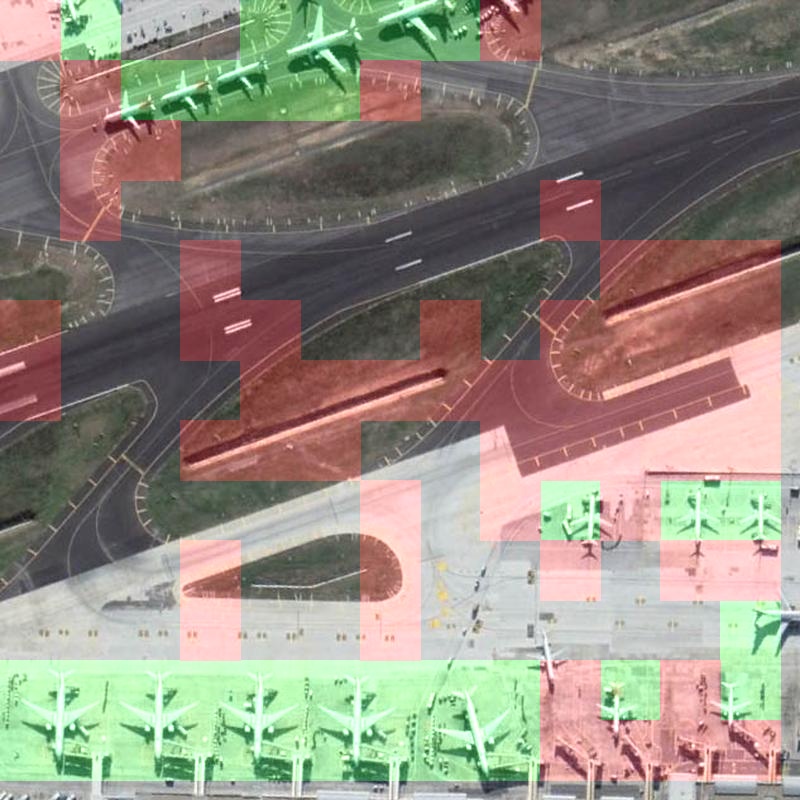}
		\caption{ }
		\label{fig:DIOR_failure1}
	\end{subfigure}
	\hfil
	\begin{subfigure}{.15\textwidth}
		\includegraphics[width=\linewidth]{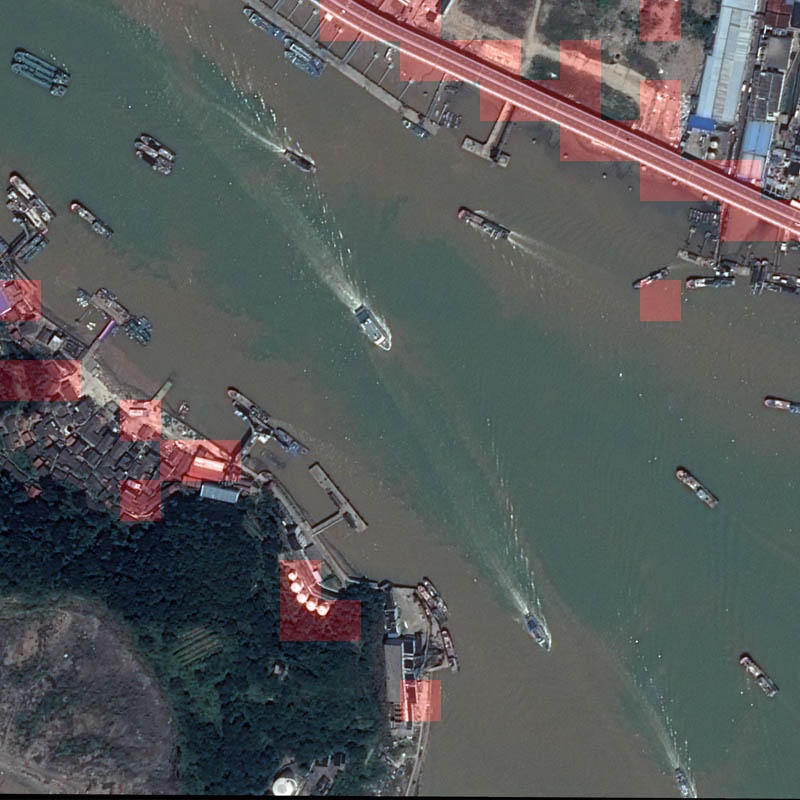}
		\caption{}
		\label{fig:DIOR_failure2}
	\end{subfigure}
	\hfil
	\begin{subfigure}{.15\textwidth}
		\includegraphics[width=\linewidth]{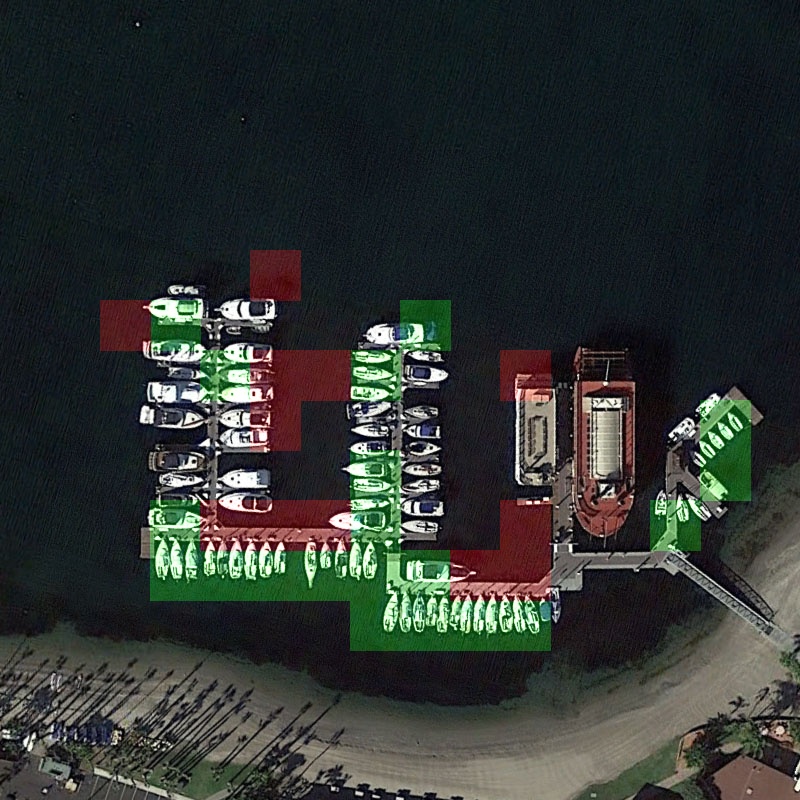}
		\caption{}
		\label{fig:DIOR_failure3}
	\end{subfigure}
	\medskip
	\begin{subfigure}{.15\textwidth}
		\includegraphics[width=\linewidth]{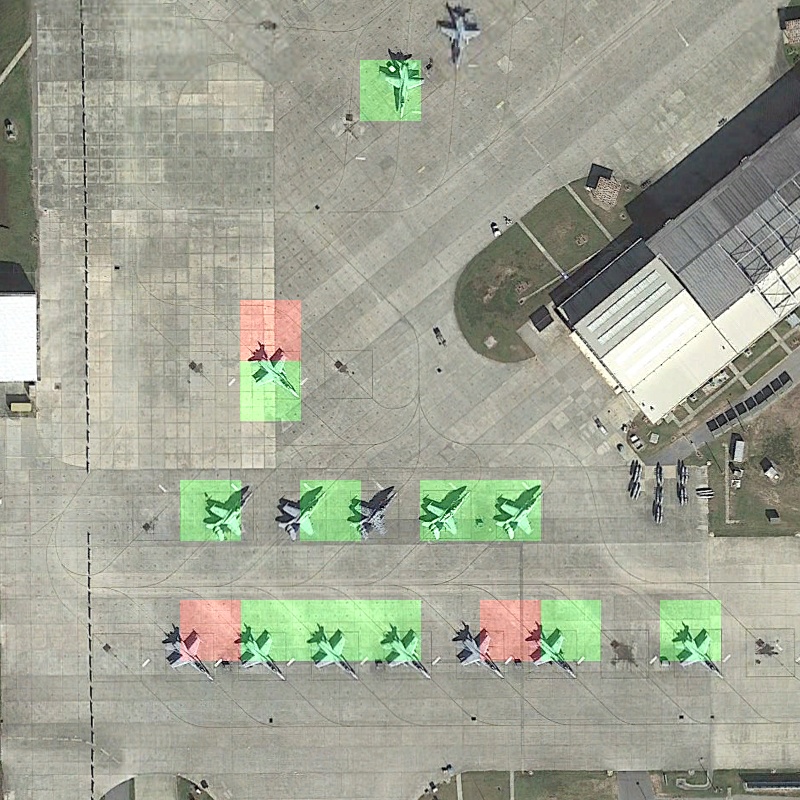}
		\caption{}
		\label{fig:DIOR_success1}
	\end{subfigure}
	\hfil
	\begin{subfigure}{.15\textwidth}
		\includegraphics[width=\linewidth]{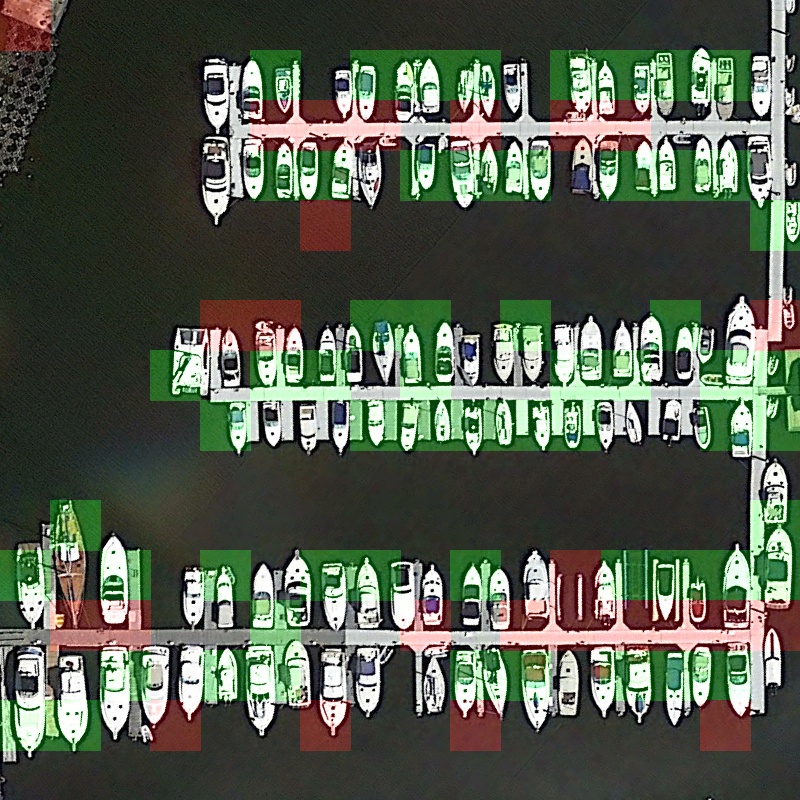}
		\caption{}
		\label{fig:DIOR_success2}
	\end{subfigure}
	\hfil
	\begin{subfigure}{.15\textwidth}
		\includegraphics[width=\linewidth]{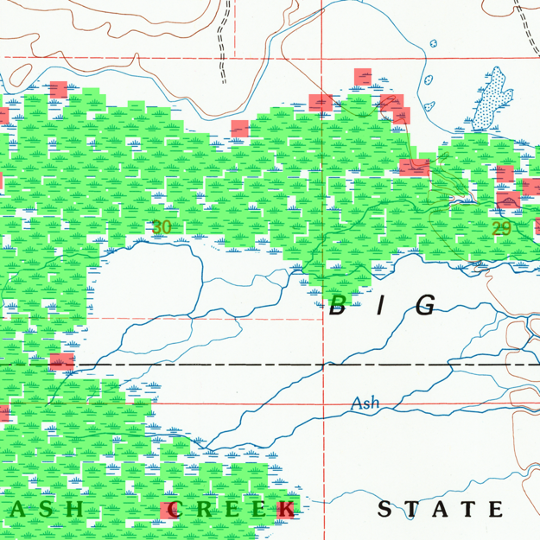}
		\caption{}
		\label{fig:USGS_success3}
	\end{subfigure}
	\caption{\small The figures in the first row show failure cases from TGGM because of ROIs and multi-scale objects. The figures in the second row show the successful cases from TGGM when ROIs are spatially-constrained and the sizes of target objects are similar.}
	\label{fig:example_TGG__failure_sucess}
\end{figure}

\subsection{Sensitivity Study}
\begin{figure}[h]
	\centering
	\includegraphics[width=.4\textwidth]{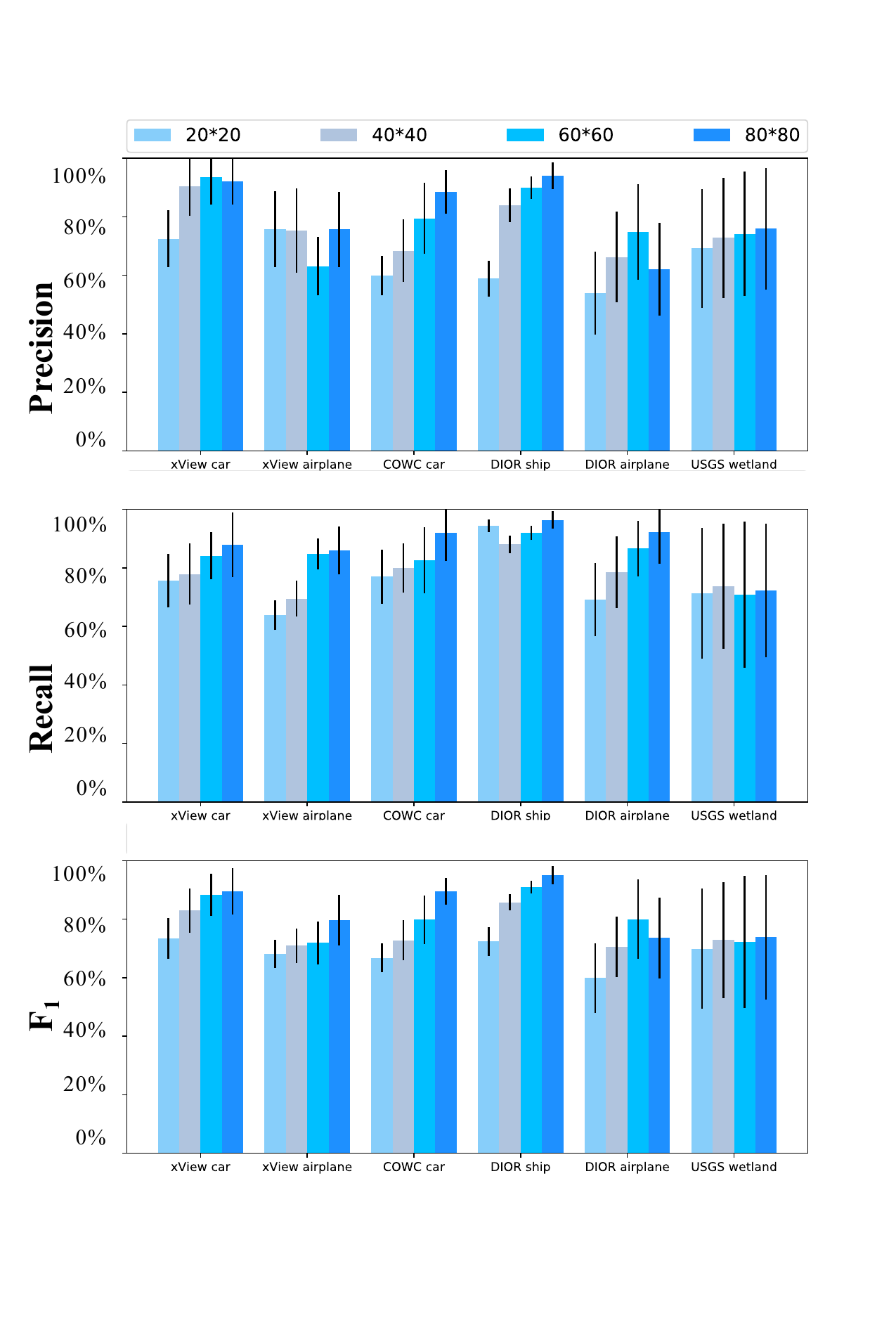}
	\caption{\small The evaluations with varied grid sizes from $20\times20$-pixel to $80\times80$-pixel. The overall precision, recall and $F_{1}$ grow with increasing grid size.}
	\label{fig:grids_chart}
\end{figure}
\subsubsection{Accuracy of Spatial Arrangement Estimation}
We varied the grid size to assess how different levels of detail affect TGGM's estimation of the spatial arrangement~\cite{grid-level_est:2}. Figure~\ref{fig:grids_chart} shows that precision, recall, and $F_{1}$ increase with increasing window sizes. In contrast, the spatial arrangement details are lost with increasing window sizes. Figure~\ref{fig:example_grid} shows an example of the spatial arrangement estimation for ships in the DIOR dataset using $20\times20$-, $40\times40$-, and $60\times60$-pixel grid dimensions, respectively. Green and red grids are true positive and false positive grids, respectively. The green and red grids in Figure~\ref{fig:example_grid} show that the performance of spatial arrangement estimation improves while the details are lost with the increase of the grid sizes. When the grid size is similar to the object size, we find that TGGM can obtain satisfactory performance of spatial arrangement estimation and acceptable spatial arrangement details. TGGM achieved around 80\% precision, recall, and $F_{1}$ on average when the grid size is similar to the object size. Figure~\ref{fig:grid_40} shows the detection results represented by a $40\times40$-pixel grid, which is similar in size to the ships in the image. The result shows that ships can be detected without losing much detail.

\begin{figure}[ht]
	\centering
	\begin{subfigure}{.15\textwidth}
		\centering
		\includegraphics[width=\linewidth]{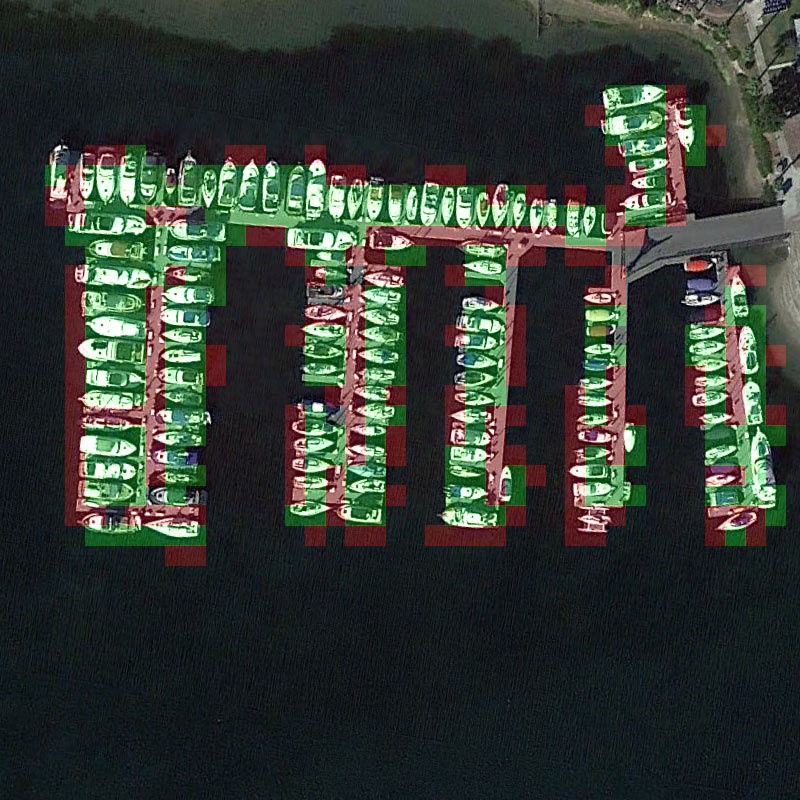}
		\caption{\small  $20\times20$ grid}
		\label{fig:grid_20}
	\end{subfigure}
	\hfil
	\centering
	\begin{subfigure}{.15\textwidth}
		\centering
		\includegraphics[width=\linewidth]{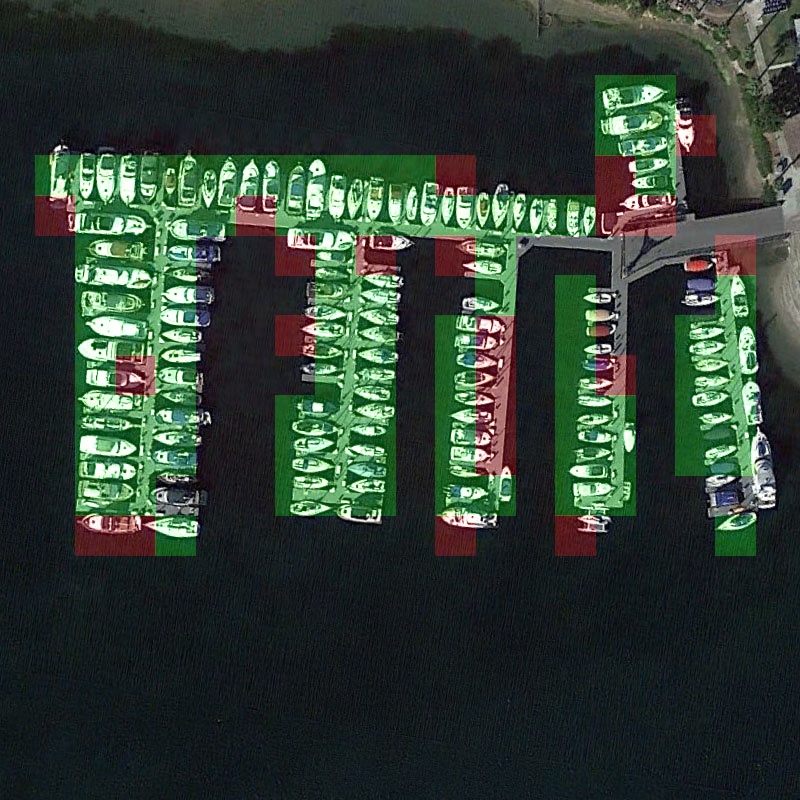}
		\caption{\small  $40\times40$ grid}
		\label{fig:grid_40}
	\end{subfigure}
	\hfil
	\centering
	\begin{subfigure}{.15\textwidth}
		\centering
		\includegraphics[width=\linewidth]{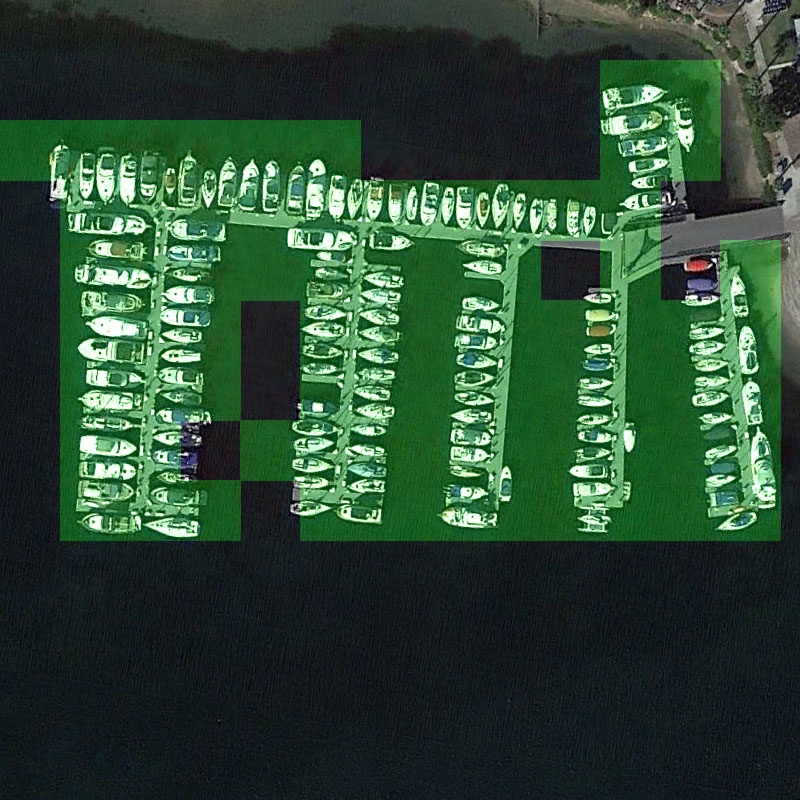}
		\caption{\small $60\times60$ grid}
		\label{fig:grid_60}
	\end{subfigure}
	\caption{\small An example of the accuracy of spatial arrangement estimation using varied grid sizes. }
	\label{fig:example_grid}
\end{figure}

\subsubsection{Spatial Arrangement Estimation over Iterations}
This group of experiments tested the robustness of TGGM to the noise in the detected target windows by showing the performance of spatial arrangement estimation over iterations. Figure~\ref{fig:iteration_chart} shows the grid-level evaluation over iterations using grid extents similar to the object sizes. All detection tasks show high precision and low recall in the first few iterations. The true positive (green) grids in Figures~\ref{fig:COWC_car_i0} and ~\ref{fig:COWC_car_i1} show that TGGM achieved precise but partial spatial arrangement estimation over the first three iterations. After the first few iterations, Figure~\ref{fig:iteration_chart} shows that precision decreases while recall increases. The green grids in Figures~\ref{fig:COWC_car_i2} and ~\ref{fig:COWC_car_i3} show gradually increasing coverage, while the red grids in Figures~\ref{fig:COWC_car_i2} and ~\ref{fig:COWC_car_i3} show the estimation contains some noise, such as partial car bodies. The detection evolvement in Figure~\ref{fig:example_iterations} shows that the TGGM at first can detect cars with similar appearances. For example, most detected cars in the first three iterations have bright colors. With more iterations, the TGGM can detect cars with diverse appearances. 

\begin{figure}[h]
	\centering
	\includegraphics[width=.45\textwidth]{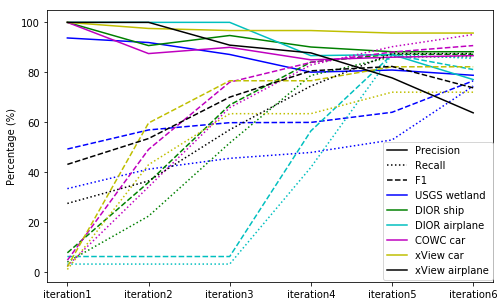}
	\caption{\small The performance evaluation over several iterations}
	\label{fig:iteration_chart}
\end{figure}

\begin{figure}[h]
	\centering
	\begin{subfigure}{.1\textwidth}
		\centering
		\includegraphics[width=\linewidth]{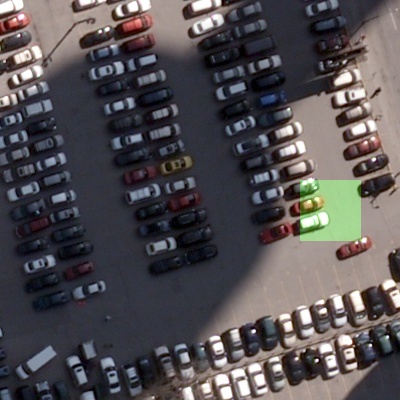}
		\caption{ \small Iteration 1}
		\label{fig:COWC_car_i0}
	\end{subfigure}
	\hfil
	\begin{subfigure}{.1\textwidth}
		\centering
		\includegraphics[width=\linewidth]{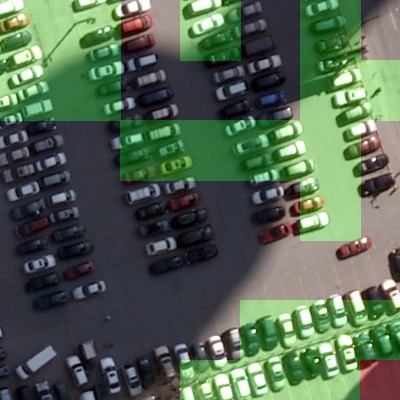}
		\caption{\small Iteration 3}
		\label{fig:COWC_car_i1}
	\end{subfigure}
	\hfil
	\begin{subfigure}{.1\textwidth}
		\centering
		\includegraphics[width=\linewidth]{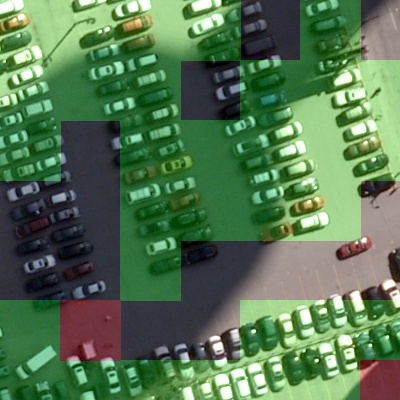}
		\caption{\small Iteration 5}
		\label{fig:COWC_car_i2}
	\end{subfigure}
	\hfil
	\begin{subfigure}{.1\textwidth}
		\centering
		\includegraphics[width=\linewidth]{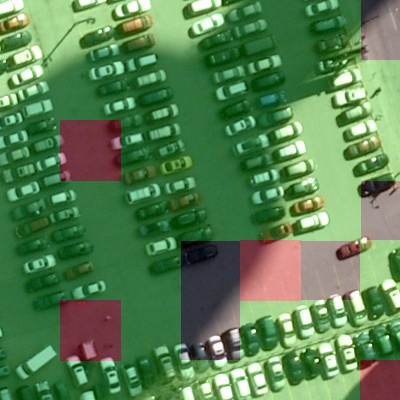}
		\caption{\small Iteration 6}
		\label{fig:COWC_car_i3}
	\end{subfigure}
	\caption{\small The spatial arrangement estimation for cars in COWC dataset using $60\times60$-pixel grids over several iterations. }
	\label{fig:example_iterations}
\end{figure}
\vspace{-10pt}
\section{Discussion}
This paper presents TGGM to estimate the spatial arrangement of the target objects within a spatially-constrained ROI in overhead images. TGGM's advantage is reducing the manual work to a few windows and an ROI annotations. The experiments show that TGGM outperforms baseline models in terms of spatial arrangement estimation accuracy and the amount of manual work. We are going to integrate TGGM into a pipeline for automatic information extraction from historical topographic maps which have little available labeled data. Historical topographic map archives store valuable information about the
evolution of natural features and human activities. Additionally, we will work to address TGGM's multi-scale objects limitation by using windows with multiple sizes.
\vspace{-2pt}
\section{Acknowledge}
This material is based upon work supported in part by the National Science Foundation under Grant Nos. IIS 1564164 (to the University of
Southern California) and IIS 1563933 (to the University of Colorado at Boulder), NVIDIA Corporation,  the National Endowment for the Humanities under Award No. HC-278125-21, and the University of Minnesota, Computer Science \& Engineering Faculty startup funds.

\vspace{-2pt}
\appendix
\section{}
Here is the detailed evidence lower bounds (ELBO) deduction for the unlabeled data $\boldsymbol{x}_{u}$ and labeled target data $\boldsymbol{x}_{t}$.

 \begin{flalign}\label{eq:detail_elbo_u}
  &\mathcal{L}_{ELBO}(\boldsymbol{x}_{u}) =     \mathbb{E}_{q(\boldsymbol{z},y|\boldsymbol{x}_{u})}\Big[\log \frac{p(\boldsymbol{x}_{u},\boldsymbol{z},y)}{q(\boldsymbol{z},y|\boldsymbol{x}_{u})}\Big]\nonumber\\
 & = \mathbb{E}_{q(\boldsymbol{z},y|\boldsymbol{x}_{u})}\Big[\log p(\boldsymbol{x}_{u}|\boldsymbol{z},y) + \log p(\boldsymbol{z}|y) + \log p(y)\nonumber\\
 &\;\;\;~ - \log q(\boldsymbol{z}|\boldsymbol{x}_{u},y) - \log q(y|\boldsymbol{x}_{u})\Big]\nonumber\\
 & = \mathbb{E}_{q(\boldsymbol{z},y|\boldsymbol{x}_{u})}\Big[\log p(\boldsymbol{x}_{u}|\boldsymbol{z},y)\Big]\nonumber\\
 &\;\;\;~ + \mathbb{E}_{q(\boldsymbol{z},y|\boldsymbol{x}_{u})}\Big[\log p(\boldsymbol{z}|y) - \log q(\boldsymbol{z}|\boldsymbol{x}_{u},y)\Big] \nonumber \\
 &\;\;\;~ + \mathbb{E}_{q(\boldsymbol{z},y|\boldsymbol{x}_{u})}\Big[\log p(y) - \log q(y|\boldsymbol{x}_{u})\Big]\nonumber\\
  &= \sum_{y}q(y|\boldsymbol{x}_{u})\int_{\boldsymbol{z}} q(\boldsymbol{z}|\boldsymbol{x_{u}},y)\Big[\log p(\boldsymbol{x}_{u}|\boldsymbol{z},y)\Big]d\boldsymbol{z}  \nonumber\\
 & - \sum_{y}q(y|\boldsymbol{x}_{u})\int_{\boldsymbol{z}}q(\boldsymbol{z}|\boldsymbol{x}_{u},y)\Big[\log q(\boldsymbol{z}|\boldsymbol{x}_{u},y) - \log p(\boldsymbol{z}|y)\Big]d\boldsymbol{z}\nonumber\\
 & - \sum_{y}q(y|\boldsymbol{x}_{u})\Big[\log q(y|\boldsymbol{x_{u}})- \log p(y)\Big] \nonumber\\
 & = \sum_{y}q(y|\boldsymbol{x}_{u})\int_{\boldsymbol{z}} q(\boldsymbol{z}|\boldsymbol{x_{u}},y)\Big[\log p(\boldsymbol{x}_{u}|\boldsymbol{z},y)\Big]d\boldsymbol{z}  \nonumber\\
 & -\sum_{y}q(y|\boldsymbol{x}_{u})\mathcal{KL}\Big[q(\boldsymbol{z}|\boldsymbol{x}_{u},y)\|p(\boldsymbol{z}|y)\Big]  -\mathcal{KL}\Big[q(y|\boldsymbol{x}_{u})\|p(y)\Big]
\end{flalign}

\begin{flalign} \label{eq:detail_elbo_l}
     &\mathcal{L}_{ELBO} (\boldsymbol{x}_{t}) =  \mathbb{E}_{q(\boldsymbol{z}|\boldsymbol{x}_{t},y=1)}\Big[\log \frac{p(\boldsymbol{x}_{t},\boldsymbol{z},y=1)}{q(\boldsymbol{z}|\boldsymbol{x}_{t},y=1)}\Big] \nonumber\\
    & = \mathbb{E}_{q(\boldsymbol{z}|\boldsymbol{x}_{t},y=1)}\Big[\log p(\boldsymbol{x}_{t}|\boldsymbol{z},y=1) + \log p(\boldsymbol{z}|y=1)\nonumber\\
     &\;\;\;~ - q(\boldsymbol{z}|\boldsymbol{x}_{t},y=1)\Big] \nonumber\\
    & = \mathbb{E}_{q(\boldsymbol{z}|\boldsymbol{x}_{t},y=1)}\Big[\log p(\boldsymbol{x}_{t}|\boldsymbol{z},y=1)\Big] \nonumber\\
    &\;\;\;~ + \mathbb{E}_{q(\boldsymbol{z}|\boldsymbol{x}_{t},y=1)}\Big[\log p(\boldsymbol{z}|y=1) - \log q(\boldsymbol{z}|\boldsymbol{x}_{t},y=1)\Big] \nonumber\\
    & = \mathbb{E}_{q(\boldsymbol{z}|\boldsymbol{x}_{t},y=1)}\Big[\log p(\boldsymbol{x_{t}}|\boldsymbol{z},y=1)\Big] \nonumber\\
    &\;\;\;~ - \mathcal{KL}\Big[q(\boldsymbol{z}|\boldsymbol{x}_{t},y=1)\| p(\boldsymbol{z}|y=1)\Big]
\end{flalign}


\newpage

{\small
\balance
\bibliographystyle{IEEEtran}
\bibliography{egbib}
}

\end{document}